\documentclass[11pt,a4paper]{article}
\usepackage[hyperref]{acl2020}
\usepackage{times}
\usepackage{latexsym}

\usepackage{url}

\usepackage{graphicx}
\usepackage{float}
\usepackage{amsmath}
\usepackage{amssymb}
\usepackage{multirow}
\usepackage{booktabs}
\usepackage{url}
\usepackage{array}
\usepackage{enumitem}
\usepackage{algorithm}
\usepackage{algorithmic}
\usepackage{pifont}

\usepackage{bm}
\usepackage{amsfonts}
\usepackage{siunitx}
\newcommand{\argmax}{\mathop{\mathrm{arg\,max}}}

\usepackage{microtype}

\aclfinalcopy % Uncomment this line for the final submission
\title{Improving Machine Reading Comprehension via Adversarial Training}

\author{Ziqing Yang$^{\dag}$, Yiming Cui$^{\dag\ddag}$, Wanxiang Che$^\ddag$,Ting Liu$^{\ddag}$, Shijin Wang$^\dag$$^\S$, Guoping Hu$^\dag$\\
{$^\dag$State Key Laboratory of Cognitive Intelligence, iFLYTEK Research, China}\\
{$^\ddag$Research Center for Social Computing and Information Retrieval (SCIR),}\\
{Harbin Institute of Technology, Harbin, China}\\
{$^\S$iFLYTEK AI Research (Hebei), Langfang, China} \\
{$^\dag$$^\S$\tt\{zqyang5,ymcui,sjwang3,gphu\}@iflytek.com}\\
{$^\ddag$\tt \{ymcui,car,tliu\}@ir.hit.edu.cn}\\
}

\date{}

\begin{document}
\maketitle
\begin{abstract}
Adversarial training (AT) as a regularization method has proved its effectiveness in various tasks, such as image classification and text classification.
Though there are successful applications of AT in many tasks of natural language processing (NLP), the mechanism behind it is still unclear.
In this paper, we aim to apply AT on machine reading comprehension (MRC)  and study its effects from multiple perspectives.
We experiment with three different kinds of RC tasks: span-based  RC,  span-based RC with unanswerable questions and multi-choice RC.
The experimental results show that the proposed method can improve the performance significantly and universally on SQuAD1.1, SQuAD2.0 and RACE.
With virtual adversarial training (VAT),  we explore the possibility of improving the RC models with semi-supervised learning and prove that examples from a different task are also beneficial.
 We also find that AT helps little in defending against artificial adversarial examples, but AT helps the model to learn better on examples that contain more low-frequency words.
\end{abstract}

\section{Introduction}
Neural networks have achieved superior performances in many tasks in the fields of computer vision (CV) and natural language processing (NLP). However, they are not robust to certain perturbations.
\citet{DBLP:journals/corr/SzegedyZSBEGF13} found that in the image classification task, neural network model predicts different labels for the original (``clean") example and the perturbed example even the difference between the two is tiny. They call perturbed examples {\it adversarial examples}. Subsequently, \citet{Goodfellow:2015} proposed {\it adversarial training} (AT) as a regularization method to improve the robustness by training on the mixture of original examples and adversarial examples. Later, in the field of NLP, \citet{DBLP:conf/iclr/MiyatoDG17} successfully applied adversarial training and \emph{virtual adversarial training }\cite{miyato2017virtual} --- a semi-supervised learning version of adversarial training --- on the text classification task. 

Though there are some successful applications of adversarial training in NLP tasks \cite{wu-etal-2017-adversarial,yasunaga-etal-2018-robust,bekoulis-etal-2018-adversarial}, the mechanism behind adversarial training in the context of NLP is still unclear, and more investigations are required to improve our understanding of adversarial training. To take one step towards this goal,  we aim to apply adversarial training on machine reading comprehension (MRC) tasks. 

MRC is an important and popular task in NLP. In MRC, a machine is asked to read a passage and then answer the questions posed based on that passage. This task is challenging since it requires sophisticated natural language understanding. Many models have been proposed and achieved superior results \cite{kadlec-etal-2016-text,cui-etal-2017-attention,DBLP:conf/iclr/SeoKFH17,devlin2018bert}. Some authors also investigated the robustness  of RC models \cite{jia-liang-2017-adversarial,wang-bansal-2018-robust}.

In this paper our goal is to improve RC models by incorporating adversarial training and analyze its effects from multiple perspectives. 
First, to verify the generality of adversarial training, we apply it to three different MRC tasks: span-based RC, span-based RC with unanswerable questions, and multi-choice RC; and conduct experiments on the representative datasets: SQuAD1.1, SQuAD2.0 and RACE.  We use BERT as our base model and adapt it to each task with task-specific modifications. The experimental results show that adversarial training consistently boosts the performance across multiple datasets.
Second, we explore the possibility of semi-supervised training on RC models with virtual adversarial training, and conclude that model can benefit from training on {\it cross-task} examples that are from other tasks. Furthermore, we investigate whether adversarial training improves the robustness of RC models on artificial adversarial examples. 
Lastly, we analyze how the model performance is improved with AT,  and find that adversarial training helps the model to  learn better on examples that contain more low-frequency words.

\section{Related Work}
{\bf Reading comprehension}. 
The objective of  MRC is to let a machine read given passages and ask it to answer the related questions. 
In recent years, more and more large-scale RC datasets became available. 
These datasets focus on different types of RC tasks, such as cloze-style RC  \cite{HermannKGEKSB15,HillBCW15}, 
span-based RC with or without unanswerable questions \cite{rajpurkar-etal-2016-squad,rajpurkar-etal-2018-know} and multi-choice RC \cite{lai-etal-2017-race}. Some tasks require the model to answer yes/no questions in addition to spans \cite{DBLP:journals/tacl/ReddyCM19}. %,DBLP:conf/emnlp/Yang0ZBCSM18}. 
With the help of the large-scale datasets, the RC models evolve rapidly and even outperform humans on some tasks \cite{cui-etal-2017-attention,DBLP:conf/iclr/SeoKFH17,DBLP:conf/iclr/XiongZS18,gpt,rmr}. However, this does not imply that machine has acquired real intelligence, as the machine can be fooled easily on artificial examples \cite{jia-liang-2017-adversarial}.

BERT \cite{devlin2018bert} as a model of pre-trained deep bidirectional representations has shown excellent performance and set the new state of the arts on various NLP tasks,  including MRC. 
It has become an indispensable part of modern high-performance RC model.
In this work, we use BERT as our base model and adapt it for different RC tasks.

{\bf Adversarial Training}.
 \citet{DBLP:journals/corr/SzegedyZSBEGF13} first discovered the existence of small perturbations to the input images that mislead models to predict wrong labels in the image classification.  They called the perturbed inputs {\em adversarial examples}.  \citet{Goodfellow:2015} proposed a simple and fast {\em adversarial training} method to improve the robustness of the model by training on both clean examples and adversarial examples.  
In the context of NLP, \citet{DBLP:conf/iclr/MiyatoDG17} applied adversarial training and virtual adversarial training \cite{miyato2017virtual} to text classification task by perturbing the word embeddings of input sentences. 
Some authors further applied  the adversarial training to various NLP tasks, such  relation extraction \cite{wu-etal-2017-adversarial} , part-of-speech tagging \cite{yasunaga-etal-2018-robust} and  jointly extracting entities and relations \cite{bekoulis-etal-2018-adversarial}.  A recent work \cite{DBLP:conf/acl/SatoSK19} investigates the effects of AT on neural machine translation.
\citet{Wang2018A3NetAN} studied the effects of applying AT to different set of variables in MRC tasks.
\citet{DBLP:conf/ijcai/SatoSS018} focuses on improving the interpretability of adversarial examples in the context of NLP.

Different from the previously discussed idea of embedding level perturbations, \citet{jia-liang-2017-adversarial} generated adversarial examples for MRC tasks at the word token level.
They introduced the {\em AddSent} algorithm, which generates adversarial examples by appending distracting sentences to the  input passages. These sentences resemble the question and do not contradict the correct answer.  They focused on the evaluation of the RC systems on these artificial adversarial examples. They showed that even the state-of-the-art RC system could be easily fooled by these adversarial examples. 
In this work, we focus on improving the generalization performance of the RC system by training on adversarial examples. 

\section{Methodology}
We first give the formal definitions of the tasks and introduce the corresponding model of each RC tasks in question. 
Then we describe the adversarial training method. Lastly we present the strategies that are useful or worth discussing in applying the AT method.

\subsection{Task Definition}
We consider three types of reading comprehension tasks: span-based extractive RC (SE-RC), span-based extractive RC with unanswerable questions (SEU-RC) and multi-choice RC (MC-RC). All of these tasks require the machine to answer questions related to the given passages.  
We denote the tokenized passage as $P=\{p_1,p_2,\ldots,p_m\}$ and the tokenized question as $Q=\{q_1,q_2,\ldots, q_n\}$. For simplicity, we use the term {\em token} and {\em word} interchangeably in the following. 

\begin{itemize}
\item {\bf SE-RC}.  Given $P$ and $Q$, the answer is a continuous span extracted from the passage : 
\begin{equation}\label{eq:validspan}
A=\{p_i,\ldots, p_j \},\ \textrm{where }1\leq i \leq j \leq m
\end{equation}
RC models on this task predict the start position $i$ and the end position $j$ of the answer.
% instead of generating the whole answer word by word.

\item {\bf SEU-RC}. Similar to SE-RC, but the passages may contain no answers. 
The model has to tell if the question is answerable, and if answerable predict the correct answer. 
Formally, for each $P$ and $Q$, the answer $A$ is either a valid span as in \eqref{eq:validspan} or empty: $A=\{\}$.

\item {\bf MC-RC}. Besides $P$ and $Q$,   additional answer options $\mathcal{O}=\{O^{(1)},O^{(2)},\ldots, O^{(m)}\}$  are provided, where each option is s sequence of words 
 $O^{(i)}=\{o^{(i)}_1, o^{(i)}_2,\ldots \}$. The model is asked to select the correct answer from $\mathcal{O}$.
\end{itemize}

\subsection{RC Model Architecture}
\begin{figure}[tbp]
\centering
\includegraphics[width=\columnwidth]{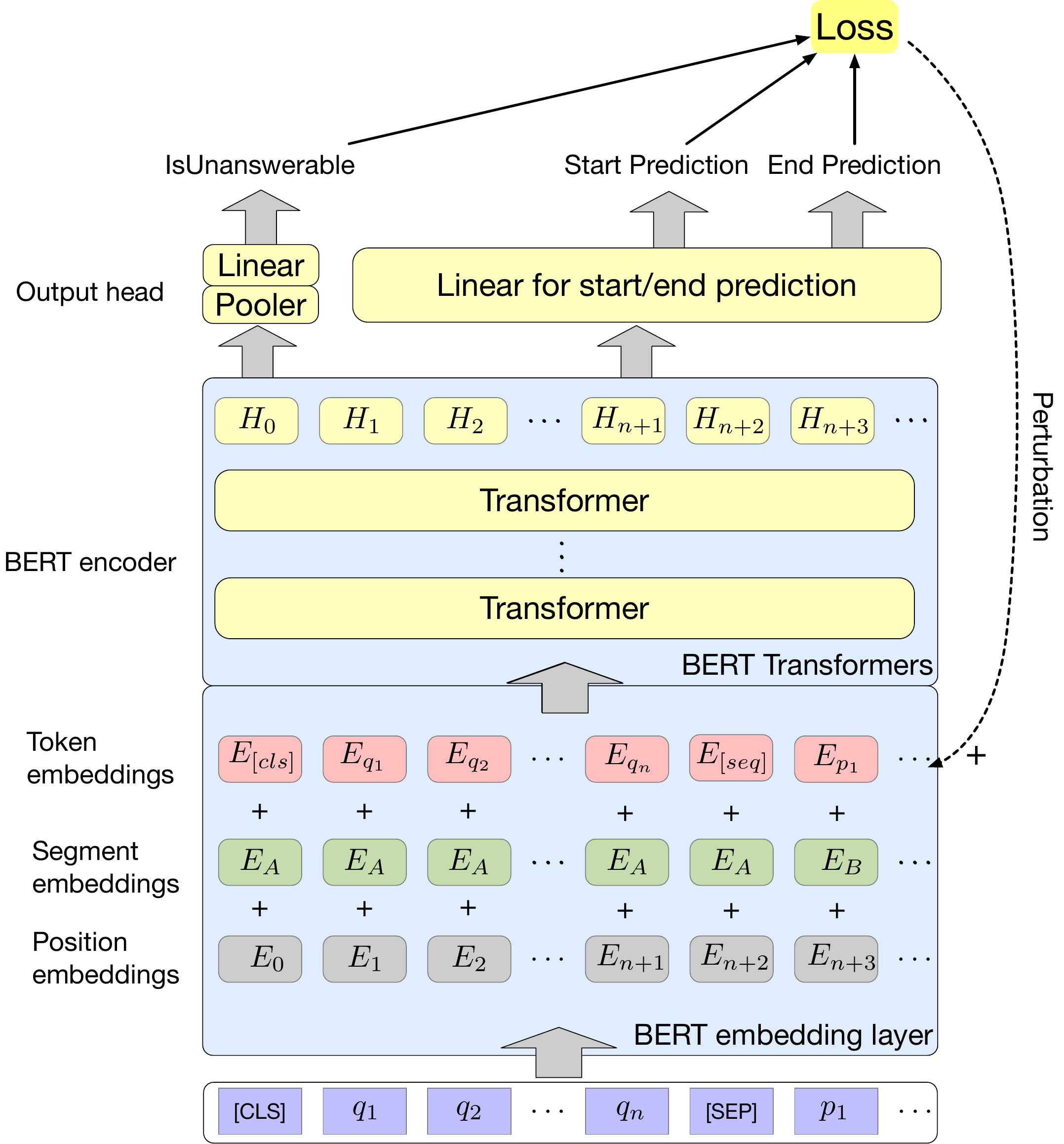}
\caption{Architecture of the model for SEU-RC. This architecture is also used in the SE-RC by simply ignoring the {\it IsUnanswerable} prediction head.  In the adversarial training phase, The perturbations are added only to the token embeddings of BERT embedding layer.}
\label{fig:model}
\end{figure}

Bidirectional Encoder Representation from Transformers (BERT)  has shown great performance and set the new state-of-the-art of various NLP tasks \cite{devlin2018bert}. It consists of an embedding layer, followed by multi-layer transformers \cite{vaswani2017attention}, and task-specific output heads. For details, the readers may refer to the original paper.
We adopt the fine-tuned BERT as our base model and adapt it to the above-listed RC tasks by designing task-specific  output heads and loss functions. 
%We describe the details below. 
The model architecture is illustrated in Figure \ref{fig:model}.

\subsubsection*{$\bullet$~~ SE-RC}
\citet{devlin2018bert} has shown how to adapt BERT to this task. 
%We briefly recap the architecture here. 
We briefly recap the input, the outputs and the loss function here.
The input is the concatenation of $P$ and $Q$ with special tokens {\tt [CLS]} and {\tt [SEP]} as {\tt [CLS]$Q$[SEP]$P$[SEP]}.
The outputs are the start/end position probability distributions $p_s$ and $p_e$.  The training objective $\mathcal{L}_{span}$ is the sum of the negative log-likelihood of  the correct start and end positions:
\begin{align}
\mathcal{L}_{span} &= \frac{1}{N}\sum_k \mathcal{L}^{(k)}_{span} \\
\mathcal{L}^{(k)}_{span}& = -y^{(k)}_s \log p^{(k)}_s - y^{(k)}_e \log p^{(k)}_e
\end{align}
where the superscript $k$ indicates the $k$-th example,  $y_s$ and $y_e$ are the ground truth start and end positions in one-hot representation, $N$ is the total number of examples.

\subsubsection*{$\bullet$~~ SEU-RC} 
The input representation and the span prediction head are the same as span-based extractive RC. We focus on how to deal with no-answer prediction. 

Some previous works introduced a special {\tt NoAnswer} token and concatenated it to the passage, or use the existed {\tt [CLS]} as the {\tt NoAnswer} token in BERT-based model \cite{unet,san2}. The no-answer prediction could be easily handled by treating {\tt NoAnswer} as a valid span position.
While in this paper, we separate the predictions for no-answer and span since it allows more flexibility and achieves a  better performance.

Denote the output from of the last transformer in BERT as $H\in \mathbb{R}^{l\times h}$, where $l$ is the sequence length and $h$ is the hidden dimension.
We use the {\em pooler} of the original BERT to squeeze $H$ into a vector $B\in \mathbb{R}^{h}$. 
 The no-answer probability is computed as
\begin{equation}
p_{na}=\mathbf{sigmoid}(B\cdot W^{na} + b^{na})
\end{equation}
where $W^{na}\in \mathbb{R}^{h}$, $b^{na}\in \mathbb{R}$. The no-answer loss of the $k$-th example is
\begin{equation}
\begin{split}
\mathcal{L}_{na}^{(k)}=&-y^{(k)}_{na} \log p^{(k)}_{na} \\
& - (1-y^{(k)}_{na})\log (1- p^{(k)}_{na})
\end{split}
\end{equation}
where $y^{(k)}_{na}=1$ if the $k$-th example is unanswerable else $0$. The total loss is  
\begin{align}\label{eq:span-na-loss}
\mathcal{L} = \frac{1}{N}\sum_k (\mathcal{L}^{(k)}_{na} + \mathcal{L}^{(k)}_{span}\cdot y^{(k)}_{na})
\end{align}
In the inference phase, we first find the most-probable valid non-empty span $A'$ and the corresponding span probability $p_{s,i}\cdot p_{e,j}$,
and compare the difference between the no-answer probability $p_{na}$ and the total span probability  $p_{s,i}\cdot p_{e,j}\cdot (1-p_{na})^2$ with a threshold (needs to be searched) to judge whether the example is answerable.

\subsubsection*{$\bullet$~~ MC-RC} 
Given  $P$,  $Q$ and $m$ options $\{O^{(1)},\ldots, O^{(m)}\}$, we construct $m$ input sequences 
\begin{center}
$X^{(1)}$={\tt [CLS]$P$[SEP]$Q$[SEP]$O^{(1)}$[SEP]} \\
\ldots \\
$X^{(m)}$={\tt [CLS]$P$[SEP]$Q$[SEP]$O^{(m)}$[SEP]}
\end{center}
and add different segmentation embedding before and after (including) $Q$. 
We feed the $m$ sequences into BERT and collect the outputs from  BERT pooler:  
\begin{equation}
{\tilde H}=\{B^{(1)},\ldots,B^{(m)}\}\in\mathbb{R}^{m\times h}
\end{equation}
The final prediction is obtained by applying a linear transformation followed by softmax over the $m$ options of ${\tilde H}$:
\begin{equation}
p_o = \mathbf{softmax}( {\tilde H}\cdot W^o + b^o)\in \mathbb{R}^{m}
\end{equation}
where $W^o\in \mathbb{R}^{h \times 1}$, $b^o\in \mathbb{R}$. The loss function is the crossentropy loss.

\subsection{Adversarial Training Method}\label{sec:at}
{\em Adversarial training} (AT) \cite{Goodfellow:2015} as a regularization method improves not only the robustness of the classifier against the perturbations but also the performance on clean inputs.  
In AT,  we first construct adversarial examples by generating worst-case perturbations that maximize the current loss function, then train the model on both of clean examples and adversarial examples.
\subsubsection*{$\bullet$~~ Adversarial Training}
In the context of MRC tasks, the inputs are sequences of words. %Since AT generates continuous perturbations,
Following \cite{DBLP:conf/iclr/MiyatoDG17}, we define the perturbation at the level of  word embeddings.
Let $\theta$ be the trainable parameters of the model. We denote the  word embedding vectors of the input sequence $X$ as 
\begin{equation}
{\bm x}=[{\bm x}_0, {\bm x}_1,\ldots,{\bm x}_{l-1}]\in \mathbb{R}^{l\times h}
\end{equation}
In our model, $\bm x$ is the token embeddings of BERT's embedding layer (see Figure \ref{fig:model}). Let $y$ denote the target. In SE-RC, $y=(y_s,y_e)$; in SEU-RC, $y=(y_s,y_e,y_{na})$; in MC-RC, $y$ is a single value representing the correct option. The worst-case perturbation $\bm{r}_{\textrm{AT}}$ is the one that maximizes the loss with a bounded norm
\begin{equation}
\bm{r}_{\textrm{AT}} = \argmax_{\bm{r};||\bm{r}||<\epsilon} \mathcal{L}({\bm x}+\bm{r};y;\hat\theta)
\end{equation}
where $\hat\theta$ means treating $\theta$ as constant.
However, the exact value of $\bm{r}_{\textrm{AT}}$ is intractable. We resort to approximating $\bm{r}_{\textrm{AT}}$  by linearizing $\mathcal{L}({\bm x}+\bm{r};y;\hat\theta)$ around $\bm x$ \cite{Goodfellow:2015} :
\begin{equation}
\bm{r}_{\textrm{AT}}=\epsilon\frac{\bm g}{||{\bm g}||}\ \textrm{, where}\ {\bm g}= \nabla_{\bm x}\mathcal{L}({\bm x};y;\hat\theta)
\end{equation}
The adversarial example is constructed as \footnote{We do not normalize word embeddings as in \cite{DBLP:conf/iclr/MiyatoDG17} and \cite{yasunaga-etal-2018-robust}, since BERT already has a LayerNorm in its embedding layer.}
\begin{equation}
{\bm x}_{\textrm {AT}}={\bm x} + \bm{r}_{\textrm{AT}} = (\hat{\bm x} + \epsilon \frac{\hat{\bm r}_{\textrm{AT}}}{||{\bm x}||})||{\bm x}||
\end{equation}
where $\hat {\bm v}$ denote the unit vector in the direction of ${\bm v}$.  
To be more clear, we redefine $\epsilon$ to be $\epsilon/||{\bm x}||$:
\begin{equation}\label{eq:xat}
{\bm x}_{\textrm {AT}}= (\hat{\bm x} + \epsilon \hat{\bm r}_{\textrm{AT}})||{\bm x}||
\end{equation}
The hyper-parameter $\epsilon$ under the new definition controls the relative strength of perturbations.

On each batch of inputs, the model is trained on both clean examples and adversarial examples by minimize the original loss $\mathcal{L}$ and the adversarial loss $\mathcal{L}_{\textrm{AT}}(\bm{x};y;\theta) = \mathcal{L}(\bm{x}_{\textrm{adv}};y;\theta)$ simultaneously:
\begin{equation}
\hat{\mathcal{L}}=\mathcal{L}(\bm{x};y;\theta)+\mathcal{L}_{\textrm{AT}}(\bm{x};y;\theta)
\end{equation}
\subsubsection*{$\bullet$~~ Virtual Adversarial Training}
{\it Virtual adversarial training} (VAT) \cite{miyato2017virtual} extends AT to semi-supervised training and unlabeled examples. VAT constructs adversarial examples by finding the perturbations that most significantly disturb the  predicted distributions:
\begin{align}
&\bm{r}_{\textrm{VAT}} = \argmax_{\bm{r};||\bm{r}||<\epsilon} \mathcal{L}_{\textrm{KL}}(\bm{x},\bm{x}+\bm{r},{\bm\theta})\\
&\mathcal{L}_{\textrm{KL}}(\bm{x},\bm{y},{\bm\theta}) = \textrm{KL}[p(\cdot|\bm{x};\hat{\bm \theta})||p(\cdot|\bm{y};{\bm \theta})]
\end{align}
where $\mathrm{KL}(\cdot||\cdot)$ is the KL divergence and $p(\cdot|\bm{x};\hat{\bm \theta})$ is the predicted distribution. The exact value of $\bm{r}_{\textrm{VAT}}$ is also intractable. An approximated solution is \cite{miyato2017virtual}
\begin{equation}
\bm{r}_{\textrm{VAT}}=\epsilon\frac{\bm g}{||{\bm g}||},\ {\bm g}= \nabla_{\xi\bm d} \mathcal{L}_{\textrm{KL}}(\bm{x},\bm{x}+\xi\bm{d},\hat{\theta})
\end{equation}
where $\bm{d}$ is a unit-norm random vector and $\xi$  is a small positive number. With the redefinition of $\epsilon$ as above, the loss of VAT is 
\begin{align}
&\mathcal{L}_{\textrm{VAT}}(\bm{x},\theta)= \mathcal{L}_{\textrm{KL}}(\bm{x},\bm{x}_{\textrm{VAT}},{\theta}) \\
&{\bm x}_{\textrm {VAT}}= (\hat{\bm x} + \epsilon \hat{\bm r}_{\textrm{VAT}})||{\bm x}||
\end{align}
VAT uses no target $y$ but only $\bm{x}$, which makes it possible to be applied on unlabeled data and perform semi-supervised training.

\subsection{Applying AT on MRC}\label{sec:strategies}
We propose the following three strategies to be used in combination with AT method in MRC tasks. We found they are either helpful, or worth discussing.
\begin{itemize}
\item {\bf  Negative entropy loss (NEL)}.
In SEU-RC, the span-prediction head is only trained on the answerable questions. 
On the unanswerable questions, we expect no spikes in the predicted span probability, which means model are less likely to make mistake. To punish any high-probability span predictions on unanswerable questions, we construct the following negative entropy loss (NEL):
\begin{align}
\mathcal{L}_{ne} &= \frac{1}{N}\sum_k(\mathcal{L}^{(k)}_{{ne}}\cdot y^{(k)}_{na}) \\
\mathcal{L}^{(k)}_{{ne}} &=\sum_i p^{(k)}_{s,i}\log p^{(k)}_{s,i} + p^{(k)}_{e,i}\log p^{(k)}_{e,i}
\end{align}
where $i$ is summed over all valid start/end positions. The intuition this is that  the uniform distribution has the highest entropy. In practice we find training $\mathcal{L}_{ne}$ only on the adversarial examples gives the best performance.

\item {\bf Semi-supervised training}.  VAT can be applied on both labeled examples and unlabeled examples. However, there are no unlabeled examples in the datasets we study. We propose to treat answerable questions as labeled examples and unanswerable questions as unlabeled examples. The details will be discussed in Section \ref{sec:ssl}.

\item {\bf Data augmentation (DA)}. We perform data augmentation to generate unanswerable questions. As will be shown in Section \ref{sec:ablation}, DA has a great influence on SQuAD2.0. The details of DA are described in Appendix A. 
\end{itemize}

\section{Experiments}
\subsection{Datasets}
We evaluate our method on the representative datasets SQuAD1.1 \cite{rajpurkar-etal-2016-squad}, SQuAD2.0 \cite{rajpurkar-etal-2018-know} and RACE \cite{lai-etal-2017-race}. 

The passages in SQuAD1.1 are retrieved from Wikipedia articles and the questions are crafted by crowd-workers.  The answer to each question is a span in the passage. The sizes of its training, development and test set are roughly 88k/11k/10k.

SQuAD2.0 contains unanswerable questions. About one-third of the examples in the training set, and half of the examples in the development and test set  are unanswerable. The sizes of its training, development and test set are roughly 130k/12k/9k.

RACE is a multi-choice RC dataset, which is collected from the English exams for middle and high school Chinese students. RACE-M denotes the middle school exams and RACE-H denotes high school exams. Each example in RACE contains four answer options, and only one of them is correct.  The sizes of its training, development and test set are roughly 88k/5k/5k.

\subsection{Experimental Setup}
\begin{itemize}[leftmargin=*,topsep=0pt,parsep=0pt]
\item {\bf BERT}. We initialize BERT with the pre-trained weights released by Google\footnote{https://github.com/google-research/bert}. 
For experiments on SQuAD, we use the {\em cased} pre-trained weights; for experiments on RACE, we use the {\em uncased} pre-trained weights.  
\item{\bf Hyper-parameters}. We set the batch size to 24, learning rate to \num{5e-5} for BERT$_{\text{\tt BASE}}$ and  \num{3e-5} for BERT$_{\text{\tt LARGE}}$. The maximum sequence length is set to $416$ for SQuAD and $512$ for RACE. The number of training epochs is $3$ for SQuAD and $5$ for RACE. We keep the other hyper-parameters of BERT  default. For adversarial training, the hyper-parameter $\epsilon$ is set to \num{1e-2} for SQuAD and \num{1e-3} for RACE.  We have found that the optimal value of $\epsilon$  of each dataset is rather stable and performs well almost in all experiments. For semi-supervised learning, batch size is 12 for unlabeled samples, and $\xi=\num{1e-5}$.
\item{\bf Evaluation}. The test sets of SQuAD1.1 and SQuAD2.0 are hidden. Thus we report the results on development sets, except the model we submitted to the official for online evaluation.
\end{itemize}
\subsection{Overall Results}\label{sec:results}
We have observed universal improvements across all three tasks, which prove the generality of adversarial training.
We first show the overall results. The analysis is provided later in Section \ref{sec:analysis}. Notice that the current state-of-the-art models (spanBERT, XLNet, RoBERTa, etc.) use different base models from BERT, which have been pre-trained from scratch on large corpora.

\begin{table}[tpb]
\resizebox{\columnwidth}{!}{
\begin{tabular}{@{}lcccc@{}}
\toprule
\multirow{2}{*}{\textbf{System}} & \multicolumn{2}{c}{\textbf{Dev}} & \multicolumn{2}{c}{\textbf{Test}} \\
                                 & EM                  & F1         & EM          & F1         \\ \midrule
\it{Human Performance}                & 80.3                & 90.5       & 82.3        & 91.2       \\ \midrule
\emph{Ensemble  model}                  &                     &            &             &            \\
nlnet$^\dag$                            & -                   & -          & 85.4        & 91.2       \\
BERT \cite{devlin2018bert}     & {86.2}                & 92.2       & {87.4}        & {93.2}       \\ \midrule
\emph{Single model}                     &                     &            &             &            \\
BERT \cite{devlin2018bert}                            & 84.2                & 91.1       & 85.1        & 91.8       \\
KT-NET\cite{yang-etal-2019-enhancing-pre}             & -                   & -          & 85.9        & 92.4       \\
BERT+WWM+MT $^\dag$    & -                   & -          & 88.7        & 94.4       \\
spanBERT \cite{DBLP:journals/corr/abs-1907-10529}         &-                    &-            & 88.8        & 94.6       \\
XLNet \cite{DBLP:journals/corr/abs-1906-08237}                 &{ 89.0}   &{ 94.5}            & {\bf 89.9}        & {\bf 95.1}       \\
RoBERTa \cite{DBLP:journals/corr/abs-1907-11692}    & {\bf 89.4}  & {\bf 94.6} &- &- \\ \hline
\it{Our single model} & & & & \\
BERT(ours)                       & 84.9                & 91.4       &-             &-            \\
{BERT+AT+VAT(2.0) }                & {86.1}                &{92.4}       & {86.9}        & {92.6}       \\
 \bottomrule
\end{tabular}}
\caption{Results on SQuAD1.1 dev/test set. Best results are in boldface. $^\dag$  indicates unpublished works. BERT(ours) is our reimplementation of BERT for SQuAD. VAT(2.0) refers to virtual adversarial training with SQuAD2.0 data. See \ref{sec:ssl} for details.} \label{table:squad1.1}
\end{table}

{\bf SQuAD1.1}.  We submitted our best single model on the development set for evaluation. The overall results are shown in Table \ref{table:squad1.1}.  Our best model BERT+AT+VAT(2.0) archives an EM/F1  score of 86.9/92.6. Compared to our BERT baseline, our model improves 1.2/1.0 on EM/F1 with $p$-value$<$0.01, which means the improvement relative to BERT is significant. Compared to the other results on the leaderboard, BERT+AT+VAT(2.0) is the best one among the BERT-based models that use weights (no whole word masking) released by Google.

{\bf SQuAD2.0}. The best model on the development set is submitted for evaluation, and the results are shown in Table \ref{table:squad2.0}.  With the help of NEL and AT,  our best model BERT+DA+NEL+AT archives 82.9/86.0 on EM/ F1. On the development set, our best model outperforms our baseline BERT+DA by 1.3/1.4  on EM/F1 respectively with $p$-value$<$0.01. 

\begin{table}[thb]
\resizebox{\columnwidth}{!}{
\begin{tabular}{llccc}
\toprule
\multirow{2}{*}{\textbf{System}}                                                         & \multicolumn{2}{c}{\textbf{Dev}} & \multicolumn{2}{c}{\textbf{Test}}                     \\
                                                                                         & EM     & F1                      & EM                        & F1                        \\ \hline
\it{Human Performance}                                                  & 86.3   & 89.0                    & 86.9                      & 89.5                      \\ \hline
\emph{Single model}                                                           &        &                         &                           &                           \\
BERT \cite{devlin2018bert}                                                                                   & 78.7   & 81.9                    & 80.0                      & 83.1                      \\
PAML+BERT$^\dag$                                                                                & -      &-   &82.6  &85.6 \\
BERT+DAE+AoA$^\dag$                                                                             & -      & -                       & { 85.9}                      & { 88.6}                      \\ 
RoBERTa \cite{DBLP:journals/corr/abs-1907-11692} &86.5 &89.4 &86.8 &89.8 \\
UPM$^\dag$  \  &- &- &87.2 &89.9 \\
XLNet + SV \cite{DBLP:journals/corr/abs-1908-05147}  &- &- &{\bf 87.2} &{\bf 90.1} \\\hline
\it{Our single model} & & & & \\
BERT + DA                                                                             & 81.5   & 84.4                    & -                         & -                         \\ 
{ BERT + DA + NEL + AT}  &82.8 &85.8 &82.9 &86.0 \\
\bottomrule
\end{tabular}}
\caption{Results on SQuAD2.0 dev/test set. Best single model results are in boldface.  $^\dag$  indicates unpublished works.  NEL refers to negative entropy loss. SV refers to SG-Net Verifier++.}
\label{table:squad2.0}
\end{table}

{\bf RACE}. Finally, we show the results on the test set of  RACE, see Table \ref{table:race}.  AT improves the overall accuracy from $66.4\%$ to $68.3\%$ on BERT$_{\text{\tt BASE}}$ and from $70.5\%$ to $72.4\%$ on BERT$_{\text{\tt Large}}$. AT method achieved significant improvements without sophisticated architecture design. 

\begin{table}[htp]
\resizebox{\columnwidth}{!}{
\begin{tabular}{@{}lccc@{}}
\toprule
\textbf{System}          & \textbf{RACE} & \textbf{RACE-M} & \textbf{RACE-H} \\ \midrule
Amazon Mechanical Turker & 73.3          & 85.1            & 69.4            \\ \midrule
{\it Single model} & & & \\

GPT \cite{gpt}                 & 59.0          & 62.9            & 57.4            \\
OCN \cite{ocn}                     & 71.7          & 76.7            & 69.6            \\
DCMN \cite{dcmn}                    & 72.3          & {77.6}            & 70.1            \\
BERT + DCMN+ \cite{dcmn} &75.8	&79.3	&74.4 \\
XLNet \cite{DBLP:journals/corr/abs-1906-08237}        &81.8	  &85.5	&80.2  \\
RoBERTa \cite{DBLP:journals/corr/abs-1907-11692}   &{\bf 83.2	}          &{\bf 86.5}	&{\bf 81.8}    \\
 \midrule
{\it Our single model} & & &  \\
BERT-base(ours)          & 66.4         & 73.5           & 63.5           \\
BERT-base + AT           & 68.3         & 73.8           & 66.1           \\
BERT-large(ours)         & 70.5         & 75.4           & 68.5           \\
BERT-large + AT         & {72.4}         & 77.0           & {70.5}           \\ \bottomrule
\end{tabular}

}
\caption{Accuracy(\%) on the test set of  RACE. We also list other competing single models on the leaderboard. }
\label{table:race}
\end{table}

\section{Analysis}\label{sec:analysis}

\subsection{Is Semi-supervised Learning Helpful?}\label{sec:ssl}
There are limited studies on semi-supervised learning on RC tasks \cite{yang-etal-2017-semi,dhingra-etal-2018-simple}. In this section, we explore this possibility with virtual adversarial training.
We conduct the experiments on SQuAD1.1, which only contains answerable questions, and treat the unanswerable questions from SQuAD2.0 training set as the unlabeled examples. We perform experiments with different configurations as shown in Table \ref{table:ssl}. 
\begin{table}[htp]
\resizebox{\columnwidth}{!}{
\begin{tabular}{@{}lccll@{}}
\toprule
\multirow{2}{*}{\textbf{System}} & \multicolumn{2}{c}{\textbf{Large}}                  & \multicolumn{2}{c}{\textbf{Base}} \\
                                 & \textbf{EM}              & \textbf{F1}              & EM              & F1              \\ \midrule
BERT                             & \multicolumn{1}{l}{84.9} & \multicolumn{1}{l}{91.4} & 81.2            & 88.7            \\
BERT+AT                          & 86.0                     & 92.2                     & 83.6            & 90.2            \\
BERT+VAT                         & 86.1                     & 92.2                     & 83.5            & 90.2            \\
BERT+AT+VAT(2.0)                & \textbf{86.1}            & \textbf{92.4}            & \textbf{83.5}   & \textbf{90.2}   \\
BERT+VAT(2.0)                    & 85.3                     & 91.9                     & 82.8            & 89.9           \\ \bottomrule
\end{tabular}
}
\caption{Results of different configurations of  AT and VAT on SQuAD1.1 development set. +AT and +VAT refers to apply AT or VAT on SQuAD1.1 training set. +VAT(2.0) refers to apply VAT on unlabeled examples from SQuAD2.0 training set.}
\label{table:ssl}
\end{table}

Training with AT or VAT solely on SQuAD1.1 training set results in similar improvements no matter on  BERT$_{\textrm{\tt BASE}}$ or BERT$_{\textrm{\tt LARGE}}$. 
Adding unanswerable questions as unlabeled examples improves performance slightly on BERT$_{\textrm{\tt LARGE}}$ (the fourth line in Table \ref{table:ssl}). 
So far, we see no significant benefits of training on unlabeled examples with labeled samples. We suppose that in order to further improve the performance with semi-supervised learning, more unlabeled examples are needed 
since typical semi-supervised learning datasets usually contains far more unlabeled examples than labeled examples \cite{DBLP:conf/iclr/MiyatoDG17}. 
However,  if we only perform VAT on unlabeled examples which is denoted as BERT+VAT(2.0) in the table,  we obtain improvements of  0.4/0.5 and 1.6/1.2 on EM/F1 relative to the baseline on  BERT$_{\textrm{\tt LARGE}}$ and BERT$_{\textrm{\tt BASE}}$ respectively. Notice that unanswerable questions are out of the domain of the SQuAD1.1 task, and the models in these experiments are not designed for handling unanswerable questions,  but with semi-supervised learning they benefit from these questions. The results prove that even cross-task data could help improving RC models.  

\subsection{Ablation Study}\label{sec:ablation}
We do an ablation study to test the effectiveness of different components in our best model BERT + DA + NEL + AT for SEU-RC on SQuAD2.0.  We run each experiment three times and report the best performance.  To further corroborate the results, we run ablation experiments on both BERT$_{\text{\tt LARGE}}$  and BERT$_{\text{\tt BASE}}$ . The ablation results are shown in Table \ref{table:ablation}.
\begin{table}[htp]
\resizebox{\columnwidth}{!}{
\begin{tabular}{@{}lccll@{}}
\toprule
\multirow{2}{*}{\textbf{System}} & \multicolumn{2}{c}{\textbf{Large}}                  & \multicolumn{2}{c}{\textbf{Base}} \\
                                 & \textbf{EM}              & \textbf{F1}              & \textbf{EM}              & \textbf{F1}              \\ \hline
BERT+ DA + NEL + AT              & 82.8                     & 85.8                     & 78.8            & 81.6            \\
-NEL                             & 82.4                     & 85.4                     & 78.6            & 81.3            \\
 -AT, -NEL                        & 81.5                     & 84.4                     & 77.2            & 80.1            \\
-DA                              & 81.0                     & 83.9                     & 76.5            & 79.1            \\
\quad -NEL                             & \multicolumn{1}{l}{80.8} & \multicolumn{1}{l}{83.7} & 76.2            & 78.9            \\
\quad -AT, -NEL                      & 78.6                     & 81.9                     & 74.2            & 76.9            \\ \bottomrule
\end{tabular}
}
\caption{Ablation study on SQuAD2.0 development set.}
\label{table:ablation}
\end{table}

Data augmentation has a critical influence on the performance, as we expected. Adversarial training boosts the performance in any configuration, no matter on BERT$_{\text{\tt LARGE}}$  or BERT$_{\text{\tt BASE}}$, with or without data augmentation, which means that adversarial training and data augmentation are two orthogonal methods. Recall that the effective number of training examples are doubled in AT as we generate adversarial examples for each input examples, thus AT can be view as a kind of data augmentation in some sense. But we see here neither the artificial data augmentation nor the automatic adversarial examples fully exploit the potential of the model by itself. Model benefits from both of them. With negative entropy loss, the performance is further improved. Though the improvement brought by NEL is not so large as AT and DA, it is stable across different configurations.

\subsection{Robustness on artificial adversarial examples}

\begin{table}[ttbp]
\resizebox{\columnwidth}{!}{
\begin{tabular}{lccccc}
\toprule
\textbf{System}  & \textbf{\begin{tabular}[c]{@{}c@{}}AddSent\\ (F1)\end{tabular}}      & \textbf{\begin{tabular}[c]{@{}c@{}}AddOneSent\\ (F1)\end{tabular}}  &  \textbf{\begin{tabular}[c]{@{}c@{}}Test\\ (F1)\end{tabular}}    & $\Delta_1$ & $\Delta_2$     \\ \hline
\emph{Single model}    &                            &                                      &                    &                   &                            \\
R.M-Reader \cite{rmr}                             & 58.5                       & 67.0                                                                         & 86.6    &  28.1    &    19.6         \\
KAR \cite{kar}                                    & 60.1                       & 72.3                                                                    & 83.5              &  23.4   &  11.2  \\
 \hline
\emph{Our single model}                                                            &                                                          &                            &          &         &         \\
BERT                              & 61.0                       & 71.1                                                                     & 91.4           & 30.4      &  20.3   \\
BERT+AT+VAT(2.0)     & 63.5                       & 72.5                   & 92.4   & 28.9 &  19.9\\  \hline
\emph{Absolute improvement(\%)} &2.5 &1.4  & 1.0 & & \\
\emph{Relative improvement(\%)} &6.4 &4.8  & 13.2 & & \\
\bottomrule
\end{tabular}}
\caption{Model performance on AddSent and AddOneSent. Results on SQuAD1.1 are also provided for comparison. $\Delta_1$ is the difference between Test(F1) and AddSent(F1); $\Delta_2$ is the difference between Test(F1) and AddOneSent(F1).}
\label{table:addsent}
\end{table}

\citet{jia-liang-2017-adversarial} constructed two artificial adversarial examples datasets called {\it AddSent} and {\it AddOneSent} based on  SQuAD1.1 by appending distracting  sentences to the passages. Models may be easily fooled on these adversarial examples and predict wrong answers from the distracting sentences because of the high overlap between the distracting sentences and the questions.  Although the generation process of these artificial adversarial examples is different from the gradient-based method used in AT, and human annotations are needed during the generation,  it is interesting to study how the AT affects the robustness of the model on these human-knowledge-injected adversarial examples.

The results are shown in Table \ref{table:addsent}.  All models are trained on SQuAD1.1 before evaluation.  Though the BERT+AT+VAT(2.0) achieves the best results on AddSent and AddSOneSent,  this is largely due to its high performance on the normal dataset, rather than obtaining additional robustness against AddSent and AddOneSent, since the relative improvements are mediocre. While KAR explicitly utilizes external general knowledge (WordNet), and it has the smallest gap $\Delta_1$ and $\Delta_2$ between F1 on test and F1 on AddSent/AddOneSent. The results show that while AT improves the generalization performance, it is not designed for defending against adversarial examples generated with human knowledge, at least in the reading comprehension tasks.  
How to bridge the gap between gradient-based and artificial adversarial examples, and how to achieve improvement and robustness on artificial adversarial examples at the same time is still an open question.

\subsection{How Does AT Help the Model Learn Better?}
AT perturbs the input directly on the embedding vectors. This operation may help to refine the word embeddings, 
especially the embeddings of low-frequency words (``rare words") since they are less trained and likely to be under-fitting.  The target task may benefit  from this refining. 
We test this hypothesis by studying the performance of the model on different groups of the examples with different number of rare words.
We sort all the words by their frequencies of occurrence in the training set and refer the last 10,000 words as rare words. We define the \emph{difficulty}
\footnote{This name is just for simplicity, not necessarily related to the true difficulty of the example. To gain some intuition on the difficulty, we show some examples in Appendix B.}
 of each example as the number of rare words in its passage and question normalized by its total number of words. We categorize all the examples in the development set by their difficulty into several buckets and study the performance on each bucket.
 
 \begin{table}[tbp]
\centering
\resizebox{\columnwidth}{!}{
\begin{tabular}{@{}lcccccc@{}}\toprule
Difficulty range  & 0$\sim$0.01 & 0.01$\sim$0.02 & 0.02$\sim$0.03 & 0.03$\sim$0.05 & \textgreater{}0.05 \\ \hline
\# of total examples    & 2676        & 2476           & 2520           & 2613           & 1588               \\
BERT+DA  & 85.4 & 85.0 & 84.1 & 83.9 & 82.4 \\
BERT+DA+AT & 86.3 & 86.1 &85.6 & 85.2 & 84.6 \\ \hline
\# of HA examples & 1356        & 1237           & 1271           & 1283           & 781                \\
BERT+DA  & 82.5 & 80.5 & 80.8 & 80.8 & 82.5 \\
BERT+DA+NEL+AT & 83.8 & 81.6 &82.5 & 81.6 & 84.4 \\ \hline
\# of NA examples & 1320        & 1239           & 1249           & 1330           & 807      \\
BERT+DA  & 88.4 & 89.5 & 87.4 & 86.9 & 82.3 \\
BERT+DA+AT & 88.9 & 90.7 &88.7 & 88.7 & 84.9 \\ \bottomrule         
\end{tabular}
}
\caption{Statistics and performance (F1) of each bucket. HA stands for answerable questions; NA stands for unanswerable questions.}
\label{table:bucket}
\end{table}

\begin{figure}[tbp]
\centering
\includegraphics[width=\columnwidth]{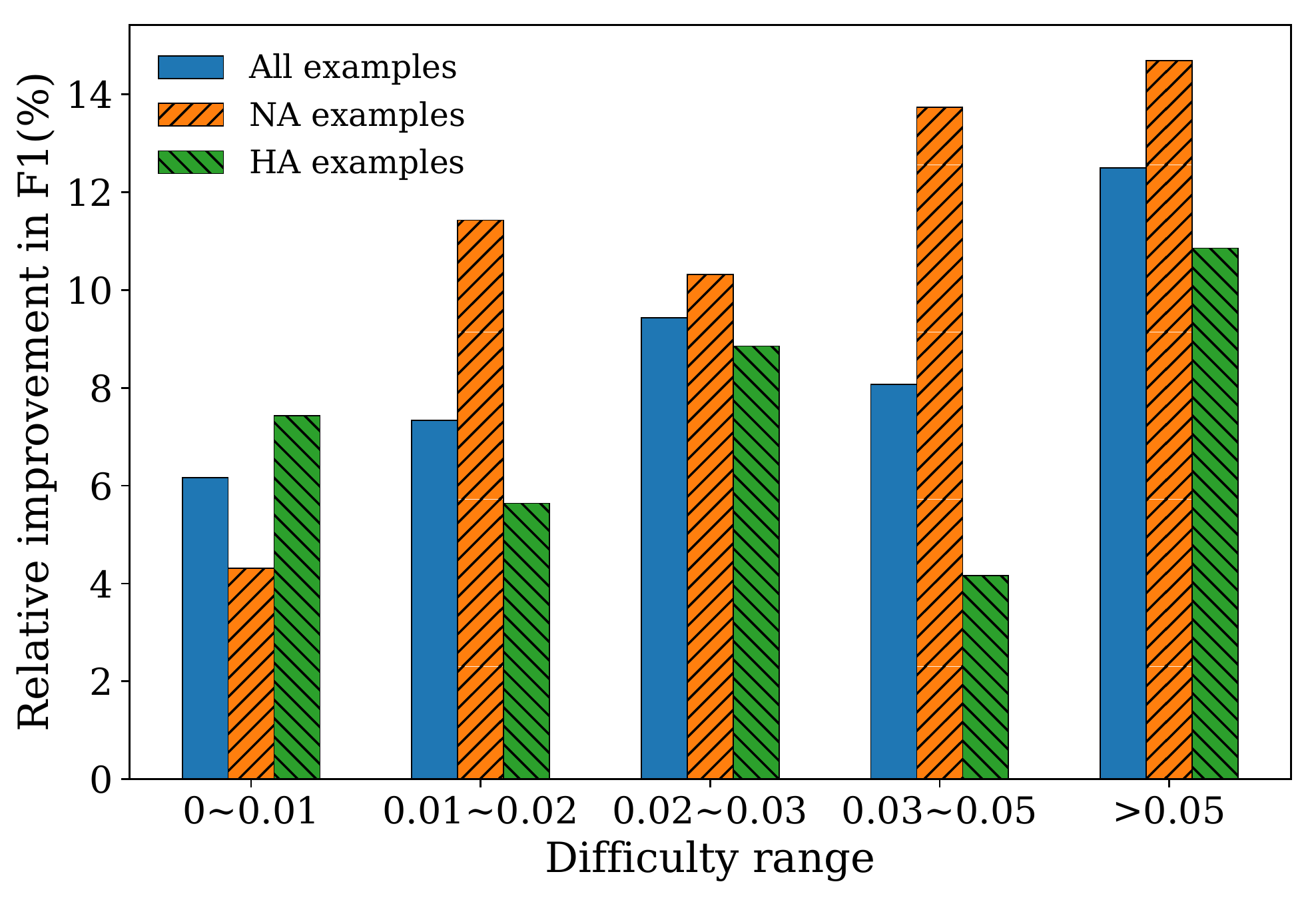}
\caption{Relative improvement of adversarial training  to the baseline model on different bucket. }
\label{fig:F1}
\end{figure}

We perform the analysis on SQuAD2.0 dataset for its variety, 
and train three BERT$_{\text{\tt LARGE}}$ baseline models and three BERT$_{\text{\tt LARGE}}$  with AT. For each group of models, we average their scores to improve stability. The results on each bucket are shown in Table \ref{table:bucket}. We plot the relative improvements on each bucket in Figure \ref{fig:F1}.  AT achieves larger improvements on more difficult examples, and the largest improvement is on the examples with difficulty$>$0.05. The increase in the relative improvement is more prominent on unanswerable (no-answer, NA) examples than answerable (has-answer, HA) examples. The reason may be that to judge whether a question is unanswerable requires the model to investigate each word in the passage so to make sure that it does not miss any important information, while a span prediction could be made by simply focusing on the context of the matching words, which means predictions on NA examples are more sensitive to the existence of rare words.

\section{Conclusion}
In this work, we applied adversarial training on MRC tasks and inspect the effects from multiple perspectives. We found that AT improves the performance significantly and consistently across different RC tasks. 
By the virtue of VAT, we performed semi-supervised learning on MRC tasks. The results show that under semi-supervised learning, the model that is not able to tackle unanswerable questions can benefit from training on unanswerable questions. This inspires us to further explore the possibility of semi-supervised learning on RC in the future.  
We also found that AT cannot defend against artificial adversarial examples.
Lastly, by a careful analysis of the effect of adversarial training on different sets of examples, we found that AT helps the model to learn better on the examples with more rare words. 

\bibliography{acl2020}

\begin{thebibliography}{38}
\expandafter\ifx\csname natexlab\endcsname\relax\def\natexlab#1{#1}\fi

\bibitem[{Bekoulis et~al.(2018)Bekoulis, Deleu, Demeester, and
  Develder}]{bekoulis-etal-2018-adversarial}
Giannis Bekoulis, Johannes Deleu, Thomas Demeester, and Chris Develder. 2018.
\newblock \href {https://www.aclweb.org/anthology/D18-1307} {Adversarial
  training for multi-context joint entity and relation extraction}.
\newblock In \emph{Proceedings of the 2018 Conference on Empirical Methods in
  Natural Language Processing}, pages 2830--2836, Brussels, Belgium.
  Association for Computational Linguistics.

\bibitem[{Cui et~al.(2017)Cui, Chen, Wei, Wang, Liu, and
  Hu}]{cui-etal-2017-attention}
Yiming Cui, Zhipeng Chen, Si~Wei, Shijin Wang, Ting Liu, and Guoping Hu. 2017.
\newblock \href {https://doi.org/10.18653/v1/P17-1055}
  {Attention-over-attention neural networks for reading comprehension}.
\newblock In \emph{Proceedings of the 55th Annual Meeting of the Association
  for Computational Linguistics (Volume 1: Long Papers)}, pages 593--602,
  Vancouver, Canada. Association for Computational Linguistics.

\bibitem[{Devlin et~al.(2018)Devlin, Chang, Lee, and
  Toutanova}]{devlin2018bert}
Jacob Devlin, Ming-Wei Chang, Kenton Lee, and Kristina Toutanova. 2018.
\newblock Bert: Pre-training of deep bidirectional transformers for language
  understanding.
\newblock \emph{arXiv preprint arXiv:1810.04805}.

\bibitem[{Dhingra et~al.(2018)Dhingra, Danish, and
  Rajagopal}]{dhingra-etal-2018-simple}
Bhuwan Dhingra, Danish Danish, and Dheeraj Rajagopal. 2018.
\newblock \href {https://doi.org/10.18653/v1/N18-2092} {Simple and effective
  semi-supervised question answering}.
\newblock In \emph{Proceedings of the 2018 Conference of the North {A}merican
  Chapter of the Association for Computational Linguistics: Human Language
  Technologies, Volume 2 (Short Papers)}, pages 582--587, New Orleans,
  Louisiana. Association for Computational Linguistics.

\bibitem[{Goodfellow et~al.(2015)Goodfellow, Shlens, and
  Szegedy}]{Goodfellow:2015}
Ian Goodfellow, Jonathon Shlens, and Christian Szegedy. 2015.
\newblock \href {http://arxiv.org/abs/1412.6572} {Explaining and harnessing
  adversarial examples}.
\newblock In \emph{International Conference on Learning Representations}.

\bibitem[{Hermann et~al.(2015)Hermann, Kocisk{\'{y}}, Grefenstette, Espeholt,
  Kay, Suleyman, and Blunsom}]{HermannKGEKSB15}
Karl~Moritz Hermann, Tom{\'{a}}s Kocisk{\'{y}}, Edward Grefenstette, Lasse
  Espeholt, Will Kay, Mustafa Suleyman, and Phil Blunsom. 2015.
\newblock \href
  {http://papers.nips.cc/paper/5945-teaching-machines-to-read-and-comprehend}
  {Teaching machines to read and comprehend}.
\newblock In \emph{Advances in Neural Information Processing Systems 28: Annual
  Conference on Neural Information Processing Systems 2015, December 7-12,
  2015, Montreal, Quebec, Canada}, pages 1693--1701.

\bibitem[{Hill et~al.(2016)Hill, Bordes, Chopra, and Weston}]{HillBCW15}
Felix Hill, Antoine Bordes, Sumit Chopra, and Jason Weston. 2016.
\newblock \href {http://arxiv.org/abs/1511.02301} {The goldilocks principle:
  Reading children's books with explicit memory representations}.
\newblock In \emph{4th International Conference on Learning Representations,
  {ICLR} 2016, San Juan, Puerto Rico, May 2-4, 2016, Conference Track
  Proceedings}.

\bibitem[{Hu et~al.(2018)Hu, Peng, Huang, Qiu, Wei, and Zhou}]{rmr}
Minghao Hu, Yuxing Peng, Zhen Huang, Xipeng Qiu, Furu Wei, and Ming Zhou. 2018.
\newblock \href {https://doi.org/10.24963/ijcai.2018/570} {Reinforced mnemonic
  reader for machine reading comprehension}.
\newblock In \emph{Proceedings of the Twenty-Seventh International Joint
  Conference on Artificial Intelligence, {IJCAI} 2018, July 13-19, 2018,
  Stockholm, Sweden.}, pages 4099--4106.

\bibitem[{Jia and Liang(2017)}]{jia-liang-2017-adversarial}
Robin Jia and Percy Liang. 2017.
\newblock \href {https://doi.org/10.18653/v1/D17-1215} {Adversarial examples
  for evaluating reading comprehension systems}.
\newblock In \emph{Proceedings of the 2017 Conference on Empirical Methods in
  Natural Language Processing}, pages 2021--2031, Copenhagen, Denmark.
  Association for Computational Linguistics.

\bibitem[{Joshi et~al.(2019)Joshi, Chen, Liu, Weld, Zettlemoyer, and
  Levy}]{DBLP:journals/corr/abs-1907-10529}
Mandar Joshi, Danqi Chen, Yinhan Liu, Daniel~S. Weld, Luke Zettlemoyer, and
  Omer Levy. 2019.
\newblock \href {http://arxiv.org/abs/1907.10529} {Spanbert: Improving
  pre-training by representing and predicting spans}.
\newblock \emph{CoRR}, abs/1907.10529.

\bibitem[{Kadlec et~al.(2016)Kadlec, Schmid, Bajgar, and
  Kleindienst}]{kadlec-etal-2016-text}
Rudolf Kadlec, Martin Schmid, Ond{\v{r}}ej Bajgar, and Jan Kleindienst. 2016.
\newblock \href {https://doi.org/10.18653/v1/P16-1086} {Text understanding with
  the attention sum reader network}.
\newblock In \emph{Proceedings of the 54th Annual Meeting of the Association
  for Computational Linguistics (Volume 1: Long Papers)}, pages 908--918,
  Berlin, Germany. Association for Computational Linguistics.

\bibitem[{Lai et~al.(2017)Lai, Xie, Liu, Yang, and Hovy}]{lai-etal-2017-race}
Guokun Lai, Qizhe Xie, Hanxiao Liu, Yiming Yang, and Eduard Hovy. 2017.
\newblock \href {https://doi.org/10.18653/v1/D17-1082} {{RACE}: Large-scale
  {R}e{A}ding comprehension dataset from examinations}.
\newblock In \emph{Proceedings of the 2017 Conference on Empirical Methods in
  Natural Language Processing}, pages 785--794, Copenhagen, Denmark.
  Association for Computational Linguistics.

\bibitem[{Liu et~al.(2018)Liu, Li, Fang, Kim, Duh, and Gao}]{san2}
Xiaodong Liu, Wei Li, Yuwei Fang, Aerin Kim, Kevin Duh, and Jianfeng Gao. 2018.
\newblock \href {http://arxiv.org/abs/1809.09194} {Stochastic answer networks
  for squad 2.0}.
\newblock \emph{CoRR}, abs/1809.09194.

\bibitem[{Liu et~al.(2019)Liu, Ott, Goyal, Du, Joshi, Chen, Levy, Lewis,
  Zettlemoyer, and Stoyanov}]{DBLP:journals/corr/abs-1907-11692}
Yinhan Liu, Myle Ott, Naman Goyal, Jingfei Du, Mandar Joshi, Danqi Chen, Omer
  Levy, Mike Lewis, Luke Zettlemoyer, and Veselin Stoyanov. 2019.
\newblock \href {http://arxiv.org/abs/1907.11692} {Roberta: {A} robustly
  optimized {BERT} pretraining approach}.
\newblock \emph{CoRR}, abs/1907.11692.

\bibitem[{Miyato et~al.(2017)Miyato, Dai, and
  Goodfellow}]{DBLP:conf/iclr/MiyatoDG17}
Takeru Miyato, Andrew~M. Dai, and Ian~J. Goodfellow. 2017.
\newblock \href {https://openreview.net/forum?id=r1X3g2\_xl} {Adversarial
  training methods for semi-supervised text classification}.
\newblock In \emph{5th International Conference on Learning Representations,
  {ICLR} 2017, Toulon, France, April 24-26, 2017, Conference Track
  Proceedings}.

\bibitem[{Miyato et~al.(2016)Miyato, Maeda, Koyama, Nakae, and
  Ishii}]{miyato2017virtual}
Takeru Miyato, Shin{-}ichi Maeda, Masanori Koyama, Ken Nakae, and Shin Ishii.
  2016.
\newblock \href {http://arxiv.org/abs/1507.00677} {Distributional smoothing by
  virtual adversarial examples}.
\newblock In \emph{4th International Conference on Learning Representations,
  {ICLR} 2016, San Juan, Puerto Rico, May 2-4, 2016, Conference Track
  Proceedings}.

\bibitem[{Radford(2018)}]{gpt}
Alec Radford. 2018.
\newblock Improving language understanding by generative pre-training.

\bibitem[{Rajpurkar et~al.(2018)Rajpurkar, Jia, and
  Liang}]{rajpurkar-etal-2018-know}
Pranav Rajpurkar, Robin Jia, and Percy Liang. 2018.
\newblock \href {https://www.aclweb.org/anthology/P18-2124} {Know what you
  don{'}t know: Unanswerable questions for {SQ}u{AD}}.
\newblock In \emph{Proceedings of the 56th Annual Meeting of the Association
  for Computational Linguistics (Volume 2: Short Papers)}, pages 784--789,
  Melbourne, Australia. Association for Computational Linguistics.

\bibitem[{Rajpurkar et~al.(2016)Rajpurkar, Zhang, Lopyrev, and
  Liang}]{rajpurkar-etal-2016-squad}
Pranav Rajpurkar, Jian Zhang, Konstantin Lopyrev, and Percy Liang. 2016.
\newblock \href {https://doi.org/10.18653/v1/D16-1264} {{SQ}u{AD}: 100,000+
  questions for machine comprehension of text}.
\newblock In \emph{Proceedings of the 2016 Conference on Empirical Methods in
  Natural Language Processing}, pages 2383--2392, Austin, Texas. Association
  for Computational Linguistics.

\bibitem[{Ran et~al.(2019)Ran, Li, Hu, and Zhou}]{ocn}
Qiu Ran, Peng Li, Weiwei Hu, and Jie Zhou. 2019.
\newblock \href {http://arxiv.org/abs/1903.03033} {Option comparison network
  for multiple-choice reading comprehension}.
\newblock \emph{CoRR}, abs/1903.03033.

\bibitem[{Reddy et~al.(2019)Reddy, Chen, and
  Manning}]{DBLP:journals/tacl/ReddyCM19}
Siva Reddy, Danqi Chen, and Christopher~D. Manning. 2019.
\newblock \href {https://transacl.org/ojs/index.php/tacl/article/view/1572}
  {Coqa: {A} conversational question answering challenge}.
\newblock \emph{{TACL}}, 7:249--266.

\bibitem[{Sato et~al.(2019)Sato, Suzuki, and Kiyono}]{DBLP:conf/acl/SatoSK19}
Motoki Sato, Jun Suzuki, and Shun Kiyono. 2019.
\newblock \href {https://www.aclweb.org/anthology/P19-1020/} {Effective
  adversarial regularization for neural machine translation}.
\newblock In \emph{Proceedings of the 57th Conference of the Association for
  Computational Linguistics, {ACL} 2019, Florence, Italy, July 28- August 2,
  2019, Volume 1: Long Papers}, pages 204--210.

\bibitem[{Sato et~al.(2018)Sato, Suzuki, Shindo, and
  Matsumoto}]{DBLP:conf/ijcai/SatoSS018}
Motoki Sato, Jun Suzuki, Hiroyuki Shindo, and Yuji Matsumoto. 2018.
\newblock \href {https://doi.org/10.24963/ijcai.2018/601} {Interpretable
  adversarial perturbation in input embedding space for text}.
\newblock In \emph{Proceedings of the Twenty-Seventh International Joint
  Conference on Artificial Intelligence, {IJCAI} 2018, July 13-19, 2018,
  Stockholm, Sweden.}, pages 4323--4330.

\bibitem[{Seo et~al.(2017)Seo, Kembhavi, Farhadi, and
  Hajishirzi}]{DBLP:conf/iclr/SeoKFH17}
Min~Joon Seo, Aniruddha Kembhavi, Ali Farhadi, and Hannaneh Hajishirzi. 2017.
\newblock \href {https://openreview.net/forum?id=HJ0UKP9ge} {Bidirectional
  attention flow for machine comprehension}.
\newblock In \emph{5th International Conference on Learning Representations,
  {ICLR} 2017, Toulon, France, April 24-26, 2017, Conference Track
  Proceedings}.

\bibitem[{Sun et~al.(2018)Sun, Li, Qiu, and Liu}]{unet}
Fu~Sun, Linyang Li, Xipeng Qiu, and Yang Liu. 2018.
\newblock \href {http://arxiv.org/abs/1810.06638} {U-net: Machine reading
  comprehension with unanswerable questions}.
\newblock \emph{CoRR}, abs/1810.06638.

\bibitem[{Szegedy et~al.(2014)Szegedy, Zaremba, Sutskever, Bruna, Erhan,
  Goodfellow, and Fergus}]{DBLP:journals/corr/SzegedyZSBEGF13}
Christian Szegedy, Wojciech Zaremba, Ilya Sutskever, Joan Bruna, Dumitru Erhan,
  Ian~J. Goodfellow, and Rob Fergus. 2014.
\newblock \href {http://arxiv.org/abs/1312.6199} {Intriguing properties of
  neural networks}.
\newblock In \emph{2nd International Conference on Learning Representations,
  {ICLR} 2014, Banff, AB, Canada, April 14-16, 2014, Conference Track
  Proceedings}.

\bibitem[{Vaswani et~al.(2017)Vaswani, Shazeer, Parmar, Uszkoreit, Jones,
  Gomez, Kaiser, and Polosukhin}]{vaswani2017attention}
Ashish Vaswani, Noam Shazeer, Niki Parmar, Jakob Uszkoreit, Llion Jones,
  Aidan~N Gomez, {\L}ukasz Kaiser, and Illia Polosukhin. 2017.
\newblock Attention is all you need.
\newblock In \emph{Advances in neural information processing systems}, pages
  5998--6008.

\bibitem[{Wang and Jiang(2018)}]{kar}
Chao Wang and Hui Jiang. 2018.
\newblock \href {http://arxiv.org/abs/1809.03449} {Exploring machine reading
  comprehension with explicit knowledge}.
\newblock \emph{CoRR}, abs/1809.03449.

\bibitem[{Wang et~al.(2018)Wang, Fu, Xu, Wu, Chen, Wei, and
  Jin}]{Wang2018A3NetAN}
Jiuniu Wang, Xingyu Fu, Guangluan Xu, Yirong Wu, Ziyan Chen, Yang Wei, and
  Lanyi Jin. 2018.
\newblock A3net: Adversarial-and-attention network for machine reading
  comprehension.
\newblock In \emph{NLPCC}.

\bibitem[{Wang and Bansal(2018)}]{wang-bansal-2018-robust}
Yicheng Wang and Mohit Bansal. 2018.
\newblock \href {https://doi.org/10.18653/v1/N18-2091} {Robust machine
  comprehension models via adversarial training}.
\newblock In \emph{Proceedings of the 2018 Conference of the North {A}merican
  Chapter of the Association for Computational Linguistics: Human Language
  Technologies, Volume 2 (Short Papers)}, pages 575--581, New Orleans,
  Louisiana. Association for Computational Linguistics.

\bibitem[{Wu et~al.(2017)Wu, Bamman, and Russell}]{wu-etal-2017-adversarial}
Yi~Wu, David Bamman, and Stuart Russell. 2017.
\newblock \href {https://doi.org/10.18653/v1/D17-1187} {Adversarial training
  for relation extraction}.
\newblock In \emph{Proceedings of the 2017 Conference on Empirical Methods in
  Natural Language Processing}, pages 1778--1783, Copenhagen, Denmark.
  Association for Computational Linguistics.

\bibitem[{Xiong et~al.(2018)Xiong, Zhong, and
  Socher}]{DBLP:conf/iclr/XiongZS18}
Caiming Xiong, Victor Zhong, and Richard Socher. 2018.
\newblock \href {https://openreview.net/forum?id=H1meywxRW} {{DCN+:} mixed
  objective and deep residual coattention for question answering}.
\newblock In \emph{6th International Conference on Learning Representations,
  {ICLR} 2018, Vancouver, BC, Canada, April 30 - May 3, 2018, Conference Track
  Proceedings}.

\bibitem[{Yang et~al.(2019{\natexlab{a}})Yang, Wang, Liu, Liu, Lyu, Wu, She,
  and Li}]{yang-etal-2019-enhancing-pre}
An~Yang, Quan Wang, Jing Liu, Kai Liu, Yajuan Lyu, Hua Wu, Qiaoqiao She, and
  Sujian Li. 2019{\natexlab{a}}.
\newblock \href {https://www.aclweb.org/anthology/P19-1226} {Enhancing
  pre-trained language representations with rich knowledge for machine reading
  comprehension}.
\newblock In \emph{Proceedings of the 57th Annual Meeting of the Association
  for Computational Linguistics}, pages 2346--2357, Florence, Italy.
  Association for Computational Linguistics.

\bibitem[{Yang et~al.(2019{\natexlab{b}})Yang, Dai, Yang, Carbonell,
  Salakhutdinov, and Le}]{DBLP:journals/corr/abs-1906-08237}
Zhilin Yang, Zihang Dai, Yiming Yang, Jaime~G. Carbonell, Ruslan Salakhutdinov,
  and Quoc~V. Le. 2019{\natexlab{b}}.
\newblock \href {http://arxiv.org/abs/1906.08237} {Xlnet: Generalized
  autoregressive pretraining for language understanding}.
\newblock \emph{CoRR}, abs/1906.08237.

\bibitem[{Yang et~al.(2017)Yang, Hu, Salakhutdinov, and
  Cohen}]{yang-etal-2017-semi}
Zhilin Yang, Junjie Hu, Ruslan Salakhutdinov, and William Cohen. 2017.
\newblock \href {https://doi.org/10.18653/v1/P17-1096} {Semi-supervised {QA}
  with generative domain-adaptive nets}.
\newblock In \emph{Proceedings of the 55th Annual Meeting of the Association
  for Computational Linguistics (Volume 1: Long Papers)}, pages 1040--1050,
  Vancouver, Canada. Association for Computational Linguistics.

\bibitem[{Yasunaga et~al.(2018)Yasunaga, Kasai, and
  Radev}]{yasunaga-etal-2018-robust}
Michihiro Yasunaga, Jungo Kasai, and Dragomir Radev. 2018.
\newblock \href {https://doi.org/10.18653/v1/N18-1089} {Robust multilingual
  part-of-speech tagging via adversarial training}.
\newblock In \emph{Proceedings of the 2018 Conference of the North {A}merican
  Chapter of the Association for Computational Linguistics: Human Language
  Technologies, Volume 1 (Long Papers)}, pages 976--986, New Orleans,
  Louisiana. Association for Computational Linguistics.

\bibitem[{Zhang et~al.(2019{\natexlab{a}})Zhang, Zhao, Wu, Zhang, Zhou, and
  Zhou}]{dcmn}
Shuailiang Zhang, Hai Zhao, Yuwei Wu, Zhuosheng Zhang, Xi~Zhou, and Xiang Zhou.
  2019{\natexlab{a}}.
\newblock \href {http://arxiv.org/abs/1901.09381} {Dual co-matching network for
  multi-choice reading comprehension}.
\newblock \emph{CoRR}, abs/1901.09381.

\bibitem[{Zhang et~al.(2019{\natexlab{b}})Zhang, Wu, Zhou, Duan, and
  Zhao}]{DBLP:journals/corr/abs-1908-05147}
Zhuosheng Zhang, Yuwei Wu, Junru Zhou, Sufeng Duan, and Hai Zhao.
  2019{\natexlab{b}}.
\newblock \href {http://arxiv.org/abs/1908.05147} {Sg-net: Syntax-guided
  machine reading comprehension}.
\newblock \emph{CoRR}, abs/1908.05147.

\end{thebibliography}
\bibliographystyle{acl_natbib}

%%%%%%%%%%%%%%%%%%%%%%%
\appendix
\section{Appendices}

\section{Data Augmentation Strategies} \label{sec:da}
We propose two simple strategies to generate unanswerable examples from the SQuAD2.0 training set.
We denote the answerable example as $(p,q,a)$,  where $p$ is the passage, $q$ is the question and $a$ is the answer. 
\subsection{Question-Passage Shuffle}
The first strategy replaces the passage $p$  in $(p,q,a)$  with another passage $p'$ that does not contain the answer text.  
Let $\mathcal{P}$ denotes all the passages that are from the same article as $p$. 
We compute the BM25 similarity score between $q$ and each passage in $\mathcal{P}$.  select the  highest-score passage $p'$ that does not contain the answer text. 
We pair $p'$ with $q$ to  generate unanswerable example $(p', q)$.

\subsection{Entity Replacement}
The second strategy generates unanswerable questions by replacing entities in the questions. Given a passage $p$,  we denote
the set of the named entities in $p$  as $\mathcal{T} = \{e_1, e_2, . . . , e_k\}$, the sets of answerable and unanswerable
questions related to $p$ as $\mathcal{Q}_a$ and $\mathcal{Q}_{na}$ respectively. For each $q$ in $\mathcal{Q}_a$,  if it contains any entity in $\mathcal{T}$,  
we generating a new question $q'$ by replacing  that entity with another randomly chosen entity in $\mathcal{T}$ that has the same entity type and does not appear in any question in $\mathcal{Q}_{na}$. The generated unanswerable example is $(p, q')$.
\newline

\noindent These two strategies generate about 70k unanswerable examples in total. We randomly choose 4k examples from question-passage shuffle and 4k examples from entity replacement as our data augmentation set.  Though the dataset is small, it is quite effective as shown in the experiments. We did not observe any significant  improvements by enlarging the data augmentation set.

\section{Examples with Different Difficulties}
To gain some intuition on the difficulty, we show some examples with different difficulties in  Table \ref{tabel:app}.
 \begin{table*}[thbp]
\begin{tabular}{p{\textwidth}}
 \toprule
{\bf Difficulty}: 0.0 \\
{\bf Passage}: {\it
After Malaysia ' s independence in 1957 , the government instructed all schools to surrender their properties and be ass \#\#im \#\#ilated into the National School system. This caused an up \#\#roa \#\#r among the Chinese and a compromise was achieved in that the schools would instead become `` National Type " schools . Under such a system , the government is only in charge of the school curriculum and teaching personnel while the lands still belonged to the schools . While Chinese primary schools were allowed to retain Chinese as the medium of instruction , Chinese secondary schools are required to change into English - medium schools . Over 60 schools converted to become National Type schools .
}\\
{\bf Question:}{\it  What language is used in Chinese primary schools in Malaysia ?} \\ 
\hline
{\bf Difficulty}: 0.017 \\
{\bf Passage}: {\it
In the triple \#\#t form , O 2 molecules are para \#\#ma \#\#gnetic . That is , they imp \#\#art magnetic character to oxygen when it is in the presence of a magnetic field , because of the spin magnetic moments of the un {\bf  \#\#pair} \#\#ed electrons in the molecule , and the negative exchange energy between neighboring O 2 molecules . Li {\bf  \#\#quid} oxygen is attracted to a magnet to a sufficient extent that , in laboratory demonstrations , a bridge of liquid oxygen may be supported against its own weight between the poles of a powerful magnet . [ c ]
}\\
{\bf Question:}{\it  What kind of field is necessary to produce a magnet effect in oxygen molecules ?} \\  \hline

{\bf Difficulty}: 0.024 \\
{\bf Passage}: {\it
According to International Mon \#\#eta \#\#ry Fund economists , inequality in wealth and income is negatively {\bf   correlated} with the duration of economic growth {\bf  spells} ( not the rate of growth ) . High levels of inequality prevent not just economic prosperity, but also the quality of a country ' s institutions and high levels of education. According to I \#\#MF staff economists , \"if the income share of the top 20 percent (the rich) increases, then GDP growth actually decline \#\#s over the medium term , suggesting that the benefits do not {\bf   trick} \#\#le down . In contrast, an increase in the income share of the bottom 20 percent (the poor) is associated with higher GDP growth . The poor and the middle class matter the most for growth via a number of interrelated economic , social , and political channels .
 }\\
{\bf Question:}{\it  What is negatively {\bf   correlated} to the duration of economic growth ?} \\  \hline

{\bf Difficulty}: 0.041 \\
{\bf Passage}: {\it
The neighborhood features restaurants , live theater and nightclub \#\#s , as well as several independent shops and bookstore \#\#s , currently operating on or near {\bf   Olive} Avenue , and all within a few hundred feet of each other . Since renewal , the Tower District has become an attractive area for restaurant and other local businesses . Today , the Tower District is also known as the center of  {\bf  Fresno} ' s LGBT and {\bf   hips} \#\#ter {\bf   Communities} . ; Additionally ,  Tower District is also known as the center of {\bf   Fresno} ' s local punk / got \#\#h / death \#\#rock and heavy metal community . [ citation needed ]
 }\\
{\bf Question:}{\it  What was Tower District known for before the renewal ?} \\  \hline

{\bf Difficulty}: 0.061 \\
{\bf Passage}: {\it
It has been argued that the term `` civil di \#\#so {\bf  \#\#bedience} " has always suffered from am \#\#bi \#\#gu \#\#ity and in modern times , become {\bf   utterly} de \#\#base \#\#d . Marshall {\bf   Cohen} notes ,  `` It has been used to describe everything from bringing a test - case in the federal courts to taking aim at a federal official . Indeed , for Vice President A \#\#gne \#\#w it has become a code - word describing the activities of {\bf  {mug}} \#\#gers , a {\bf  \#\#rson} \#\#ists , draft e \#\#vade \#\#rs, campaign {\bf   heck} \#\#lers, campus militants , anti-war demons \#\#tra \#\#tors , juvenile del \#\#in \#\#quent \#\#s and political {\bf  assassins} . " 
}\\
{\bf Question:}{\it  Vice President A \#\#gne \#\#w describes Civil di \#\#so {\bf   \#\#bedience} in what activities?} \\

  \bottomrule
\end{tabular}
\caption{Examples with different difficulties from SQuAD2.0 development set. We show  the passages and questions after tokenization. Rare words (tokens) are shown in {\bf   bold}.}
\label{tabel:app}
\end{table*}

\end{document}